# ALMRR: Anomaly Localization Mamba on Industrial Textured Surface with Feature Reconstruction and Refinement

Shichen Qu[1,2], Xian Tao[✉][1,2], Zhen Qu[1,2], Xinyi Gong[3], Zhengtao Zhang[1,2], Mukesh Prasad[4]

[1] Institute of Automation, Chinese Academy of Sciences, Beijing, China
[2] School of Artificial Intelligence, University of Chinese Academy of Sciences, Beijing, China
[3] Space Information Research Institute, Hangzhou Dianzi University, Hangzhou, China
[4] Faculty of Engineering and Information Technology, University of Technology Sydney, Sydney, Australia
qushichen23@mails.ucas.ac.cn, taoxian2013@ia.ac.cn

**Abstract.** Unsupervised anomaly localization on industrial textured images has achieved remarkable results through reconstruction-based methods, yet existing approaches based on image reconstruction and feature reconstruction each have their own shortcomings. Firstly, image-based methods tend to reconstruct both normal and anomalous regions well, which lead to over-generalization. Feature-based methods contain a large amount of distinguishable semantic information, however, its feature structure is redundant and lacks anomalous information, which leads to significant reconstruction errors. In this paper, we propose an Anomaly Localization method based on Mamba with Feature Reconstruction and Refinement(ALMRR) which reconstructs semantic features based on Mamba and then refines them through a feature refinement module. To equip the model with prior knowledge of anomalies, we enhance it by adding artificially simulated anomalies to the original images. Unlike image reconstruction or repair, the features of synthesized defects are repaired along with those of normal areas. Finally, the aligned features containing rich semantic information are fed into the refinement module to obtain the anomaly map. Extensive experiments have been conducted on the MVTec-AD-Textured dataset and other real-world industrial dataset, which has demonstrated superior performance compared to state-of-the-art (SOTA) methods.
Our code will be available at: https://github.com/qsc1103/ALMRR

## 1 Introduction

Unsupervised anomaly localization aims to precisely detect and locate abnormal regions in industrial images using prior knowledge from only anomaly-free images. Due to the strict quality control in industrial production line, samples with anomalies are extremely rare, and manual annotation can be overly time-consuming and costly[1]. This leads to a highly imbalanced training set, typically containing only images without anomalies. More importantly, various unpredictable anomalies may occur during the manufacturing process, hence detection models trained only on known anomalies may not generalize well to unseen abnormal samples. In recent years, unsupervised anomaly detection and localization has garnered widespread attention and achieved significant success.

Anomaly is an area that shows a significant deviation from the normal predefined characteristics. Based on this premise, generative models based on reconstruction[3,4,5,6,7,8] are commonly used. These models are based on the strong assumption that normal areas can be almost perfectly reconstructed through the model, whereas abnormal areas fail to reconstruct, meaning there is a significant difference between abnormal and normal areas. These methods rely on reconstruction errors to detect and locate anomalies. However, the generalization ability of reconstruction network is often underestimated, the network can not only reconstruct normal areas but can also generalize to abnormal areas which lead to missed detection of abnormal areas.

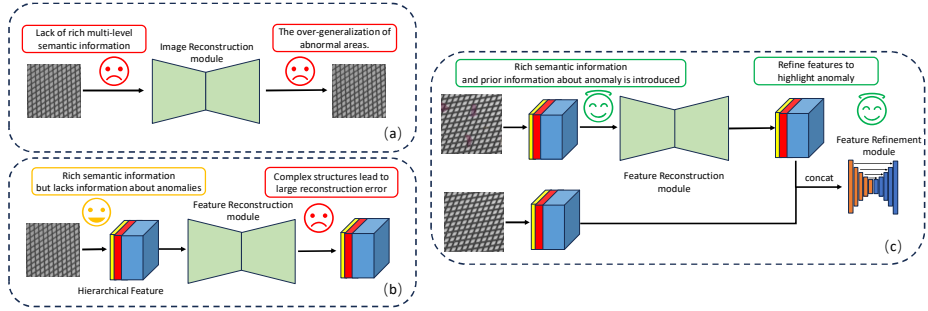

**Fig. 1.** Comparison of our method's framework with existing frameworks. (a) Image-level reconstruction lacks rich multi-level semantic information and tends to over-generalize abnormal areas. (b) Feature-level reconstruction possesses rich semantic features, but the complex structure of these features leads to significant reconstruction errors. (c) Our method: Under the framework of feature-based reconstruction, dual-branch feature extraction structure and feature refinement module are introduced. The model uses prior information about anomalies combined with feature refinement to delineate decision boundaries more clearly.

As shown in Fig.1 (a), image-based reconstruction methods reconstruct multi-channel original pixel values, and their lack of rich semantic information and excessive generalization capabilities lead to poor performance on textured images. Therefore, feature-based reconstruction methods shown in Fig.1 (b) have been proposed. The feature extraction network pretrained on a large public dataset (such as ImageNet[12]) is employed to extract rich and highly recognizable multi-scale feature information from images. However, the complexity and redundancy of feature structure and the lack of prior knowledge about anomaly, making it difficult to reconstruct accurately. The above two methods only learn the information from normal samples and lack information on anomalies, which leads to unclear decision boundaries between normal and abnormal distributions. Our proposed model, as shown in Fig.1(c), employs a multi-scale feature reconstruction framework. During the training stage, simulated anomalies with known masks are utilized to enrich the semantic understanding of anomalies. Additionally, the feature refinement module highlights anomalous regions effectively.

In the early stages, reconstruction-based anomaly detection methods primarily used Auto-Encoders(AE)[3,4], Variational Auto-Encoders(VAE)[5,6], and Generative Adversarial Networks(GAN)[7,8]. However, these methods have limited receptive fields that hinder their ability to model remotely, making them ineffective at extracting features in images. In recent years, many reconstruction-based methods[17,20,23] based on Transformer[9] have emerged. However, despite their global receptive fields, Transformers suffer from quadratic computational complexity issues, making them inefficient at multi-scale feature reconstruction tasks. Recent studies have shown that state-space models (SSM), represented by Mamba[31], can effectively capture long-range relations while maintaining linear computational complexity. We proposed a feature reconstruction module built on vision Mamba[2] framework. To our knowledge, it is the first time Mamba has been applied to feature reconstruction-based anomaly localization.

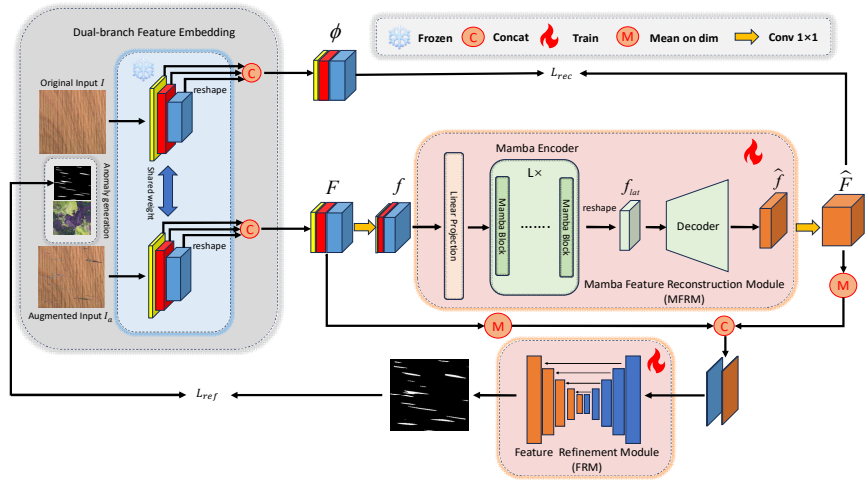

**Fig. 2.** The overview of our method's pipeline which consists four stages: simulate anomaly process, dual-branch feature embedding, Mamba feature reconstruction module and feature refinement module. Figure Best viewed in color.

In this paper, we propose a novel anomaly detection method named Anomaly Localization Mamba with Feature Reconstruction and Refinement (ALMRR). It employs a frozen pretrained CNN backbone in a dual-branch manner to extract features from both the original and augmented images. Mamba-based reconstruction module aims to repair abnormal features. The features before and after repair are fed into a feature refinement module for precise segmentation and localization of the anomaly areas. This method boasts strong feature representation and anomaly refinement capabilities. Extensive experiments confirm that our method achieves high performance on textured images and sets a new state-of-the-art for anomaly detection and localization performance on MVTec-AD-Textured[11] dataset.

The contributions of our work are as follows:

- We proposed a new research paradigm that employs a dual-branch approach for feature reconstruction and refinement. Through this method, both the accuracy of detection and the visual quality of the anomaly map have been significantly enhanced.
- To the best of our knowledge, ALMRR is the first model to introduce Mamba into feature reconstruction-based anomaly localization, reducing computational complexity while retaining a global receptive field.
- On the MVTec-AD-Textured[11] benchmark dataset, our method achieved state-of-the-art performance, and additional experiments have proven that it can be effectively applied to other real-world industrial datasets.

## 2    Related work

Unsupervised anomaly detection can only use anomaly-free samples for training, thus the most crucial task is to learn as much as possible about the features or distribution of normal samples from limited data. Currently, the mainstream methods fall into three major categories: Featuring-embedding-based methods, Reconstruction-based methods and Synthesizing-anomaly-based methods.

**Featuring-embedding-based methods.** These methods have demonstrated through extensive research that pretrained models on large datasets have strong feature representation capabilities. Featuring-embedding-based methods embed these discriminative features into a feature space and then distinguish between normal and abnormal samples based on the embedding distributions of these samples. PaDim[13] utilizes multivariate Gaussian distributions to obtain a probabilistic representation and evaluates anomaly scores via the Mahalanobis distance. PatchCore[14] establishes a memory bank of normal samples and uses greedy coreset subsampling to lighten it. Memseg[15] introduces a memory bank of normal patterns within the U-Net structure to assist the model learning.

**Reconstruction-based methods.** These methods rely on reconstruction errors for detection and localization and are divided into two main paradigm: image-based reconstruction and feature-based reconstruction. For image-based reconstruction: DAGAN [8] is developed to solve the problem of sample imbalance which exhibits excellent image reconstruction ability and training stability. DiffusionAD[21] uses the recent Diffusion model to reconstruct image more precisely. For feature-based reconstruction: DFR[16] utilizes CNN to reconstruct multi-scale pyramid features, and then identifies abnormal areas through Euclidean distance. Transformer-based framework is employed by ADTR[17] for feature reconstruction.

**Synthesizing-anomaly-based methods.** Perlin noise is used in DRAEM[18] as simulated anomaly to train the model's reconstruction and discrimination capabilities. RIAD[19] and Intra[20] utilize CNN and Transformer, respectively, to inpaint the image by recovering all masked patches. FAIR[22] proposes a novel frequency-aware self-supervised image reconstruction task that mitigates adverse generalization on anomalies. DSR[28] generates feature-level anomalies to sample the learned quantized feature space.

## 3 Method

### 3.1 General Design of ALMRR

To address the limitations in existing frameworks for anomaly detection based on feature-level and image-level reconstruction, we propose a method named ALMRR. The pipeline is shown in Fig. 2.

During training, anomaly-free images first pass through a noise generator to obtain simulated anomalies with known masks. Then, both the anomaly-free and simulated anomaly images obtain multi-scale feature extraction as $\phi \in \mathbb{R}^{C \times H \times W}$ and $F \in \mathbb{R}^{C \times H \times W}$ via a dual-branch feature embedding module with shared weights, where 1x1 convolutional layers are used for dimension reduction/expansion at the front and end of the network. The simulated anomaly dense feature blocks $f \in \mathbb{R}^{C_1 \times H \times W}$ are fed into the Mamba Feature Reconstruction Module (MFRM) for feature reconstruction using anomaly-free dense feature blocks $\phi$ as reference. Finally, the feature blocks before and after reconstruction perform mean operation over the dimension and are fed into the Feature Refinement Module (FRM) for refining anomaly localization. The simulated anomalies with known masks serve as the reference.

In the inference stage, real anomaly images, after multi-scale feature extraction, are directly fed into the reconstruction module (without the addition of simulated anomalies), and perform mean operation on the dimension, then input it into the feature refinement module to obtain the predicted anomaly map.

### 3.2 Simulate anomaly process and Dual-branch feature embedding

First, Perlin noise generator and an additional texture image $A$ is employed to mask the training images $I$ and obtain simulated anomalous images $I_a$ (shown in Fig. 3), the process that can be defined as:

$$I_a = \overline{M} \odot I + (1-\alpha)(M \odot I) + \alpha(M \odot A) \tag{1}$$

Where $I_a$ is the artificially simulated anomalous image, $I$ is the original training image, $M$ is the Perlin noise mask, $\overline{M}$ is the inverse of $M$, $\odot$ is the element-wise multiplication operation, $\alpha$ is the opacity parameter in blending, $A$ is an additional texture image.

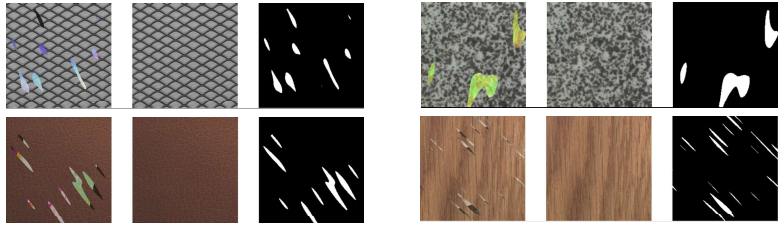

**Fig. 3.** Samples of artificially simulated anomaly

A frozen ResNet-50[10] pre-trained on ImageNet[12] is utilized as backbone. It is important to note that we adopt a dual-branch approach to perform multi-scale feature extraction on both the original $I$ and the simulated anomalous images $I_a$. The features of the original image $\phi \in \mathbb{R}^{C \times H \times W}$ serve as reference in subsequent modules for learning the distribution of normal features during reconstruction. Due to the lack of abstract information in textured industrial images, we only select the shallower features of ResNet-50 as the reconstruction target. The multi-scale features resize and concatenate together as $F \in \mathbb{R}^{C \times H \times W}$.

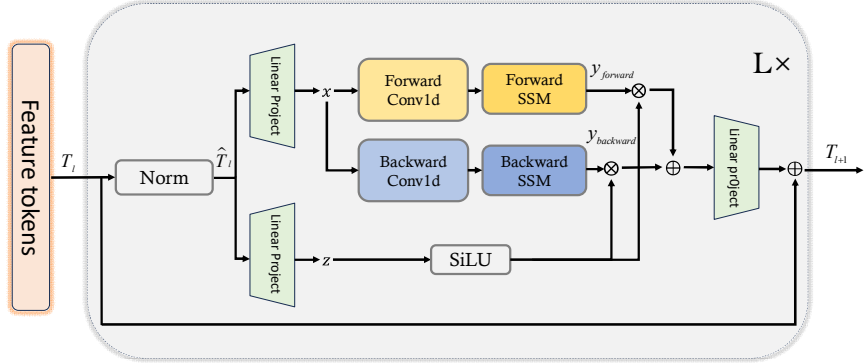

**Fig. 4.** The architecture of Mamba block

### 3.3 Mamba Feature Reconstruction Module (MFRM)

The Mamba feature reconstruction module is shown in Fig. 2, before feeding the feature token into the Mamba encoder, the feature map $F \in \mathbb{R}^{C \times H \times W}$ is first passed through a 1x1 convolutional layer to reduce the dimensionality and lessen computation consumption. Correspondingly, their dimensions are recovered by another 1×1 convolution when output by decoder.

The Mamba[2] encoder embeds the input feature tokens into a latent feature space. Each Mamba block follows the standard architecture as is shown in Fig. 4. First, the feature maps $f$ are converted into feature tokens $T_l$ using a linear projection and add position embeddings. The feature token sequence is then normalized by a normalization layer, followed by a linear projection of the sequence onto the $x$ and $z$ axis. $x$ is processed in both forward and backward directions as $y_{forward}$ and $y_{backward}$. For each direction, a 1-D convolution and an SSM module are utilized. Finally, the outputs from the forward and backward directions are gated by SiLU(z) and combined to produce the output feature tokens $T_{l+1}$. The process can be described as follows:

$$\widehat{T}_l = Norm(T_l) \tag{2}$$

$$x = Linear_x(\widehat{T}_l) \tag{3}$$
$$z = Linear_z(\widehat{T}_l)$$

$$y_{forward} = SSM_{forward}(Conv1d_{forward}(x))$$
$$y_{backward} = SSM_{backward}(Conv1d_{backward}(x)) \tag{4}$$

$$\widehat{y}_{forward} = y_{forward} \odot SiLU(z)$$
$$\widehat{y}_{backward} = y_{backward} \odot SiLU(z) \tag{5}$$

$$T_{l+1} = Linear(\widehat{y}_{forward} + \widehat{y}_{backward}) + T_l \tag{6}$$

As the goal of the decoder is to reconstruct the feature results in the original 2-D image space, the feature embeddings need to be reshaped to a feature map with the shape $f_{lat}$. The decoder is used to decode the latent feature map $f_{lat}$ to the original feature shape $\widehat{F} \in \mathbb{R}^{C \times H \times W}$. For the decoder, three transposed convolutional layers, with instance normalization and ReLU in-between are utilized, and tanh as the final nonlinearity.

MFRM is trained with a reconstruction loss measured by the averaged pair-wise $l_2$-distance between the reconstructed dense feature representation and anomaly-free feature representation extracted from the dual-branch network.

$$L_{rec} = \sum_{i=1}^{h_0} \sum_{j=1}^{w_0} \left\| f_{i,j}(x) - \widehat{f}_{i,j}(x) \right\|_2 \tag{7}$$

Note that $f(x)$ and $\widehat{f}(x)$ are feature maps with the same size $\mathbb{R}^{C_1 \times H \times W}$.

### 3.4 Feature Refinement Module (FRM)

Due to the high complexity of the multi-scale feature structure, solely relying on distance measurement could introduce reconstruction errors. Therefore, we propose a Feature Refinement Module (FRM), which adopts a structure similar to U-net. This module first takes mean operation along the dimension direction of the dense feature blocks from the output $\widehat{F}$ and input $F$ of the feature reconstruction module, and then concatenates them to serve as the input for the FRM. There are significant differences in feature distribution between $\widehat{F}$ and $F$, which provides the necessary conditions for localization. The output of this module is an anomaly score map of the same size as the input image. Generally, defective areas are smaller in size compared to normal areas. To accommodate this imbalance of positive and negative samples, focal loss combined with dice loss was employed as the loss function to enhance the accuracy of refined segmentation.

$$L_{focal} = -\alpha_t (1-p_t)^\gamma \log(p_t) \qquad (8)$$

Where $\alpha_t$ is the balancing factor, $\gamma$ is the focusing parameter, $(1-p_t)^\gamma$ is the modulating factor.

$$L_{dice} = 1 - \frac{2|M_a \cap M|}{|M_a| + |M|} \qquad (9)$$

Where $M_a$ and $M$ are the predict anomaly localization mask and ground truth mask, respectively.

The loss function for the refinement module is:

$$L_{ref} = L_{focal} + L_{dice} \qquad (10)$$

Considering both the refinement and the reconstruction loss of the two sub-module, the total loss function used in training ALMRR is:

$$L_{total} = L_{rec} + L_{ref} \qquad (11)$$

## 4 Experiments

### 4.1 Dataset and Evaluation Metric

In this work, we evaluated our proposed model using three textured datasets, including the MVTec-AD-Textured dataset[11], MT Defect dataset[24] and NanoTWICE[25]. Here is a brief introduction to these datasets:

- **MVTec-AD-Textured** contains 1781 high-resolution color images of five texture types. It includes normal images (i.e., without defects) for training purposes, as well as abnormal images for testing. There are various different types of defects in the abnormal images, such as scratch, hole, color and bent. Since its release, the dataset has served as a benchmark for unsupervised anomaly detection.
- **MT Defect** contains a total of 1344 images, with the region of interest (ROI) of the magnetic tiles cropped out. It includes 6 categories of images: normal samples and five types of defects (blowhole, crack, fray, break and uneven). To simulate the manufacturing process on a real industrial production line, images of the given magnetic tiles were captured under various lighting conditions.
- **NanoTWICE** contains 45 nanofibrous material images of size 1024×3696 pixels, of which 5 are normal, and the remaining 40 contain some form of anomaly. These anomalies are typically very small, consisting of just a few pixels.

The standard metric in anomaly detection, AUROC, is employed as the evaluation metrics throughout our study. Image-level AUROC is used for anomaly detection and pixel-based AUROC for anomaly localization. However, because the abnormal regions are very small and there is a significant imbalance between positive and negative samples, the AUROC does not effectively reflect the result of localization. Therefore, pixel-level AP is introduced to evaluate the accuracy of anomaly localization.

### 4.2 Implementation details

All input images are resized to a dimension of 256 × 256 pixels, then fed into the model. We utilize a frozen ResNet50 pre-trained on ImageNet as our backbone and reshape the feature maps to 64 × 64 pixels. Specifically, the residual connection blocks from block1 to block3 are selected to extract multi-scale dense features. Within the Mamba block, we set the hidden state dimension D to 192, the number of blocks L to 8 and use a 4×4 kernel size projection layer to get a 1-D sequence of nonoverlapping patch embeddings, then subsequently add position embedding. During training, the batch size is fixed to 4, and the training epoch was empirically set to 700. Our model is learned by Adam optimizer with the initial learning rate $1 \times 10^{-4}$. All the experiments are performed on a single NVIDIA A100-PCIE graphics processing unit(GPU) with 40 GB of memory.

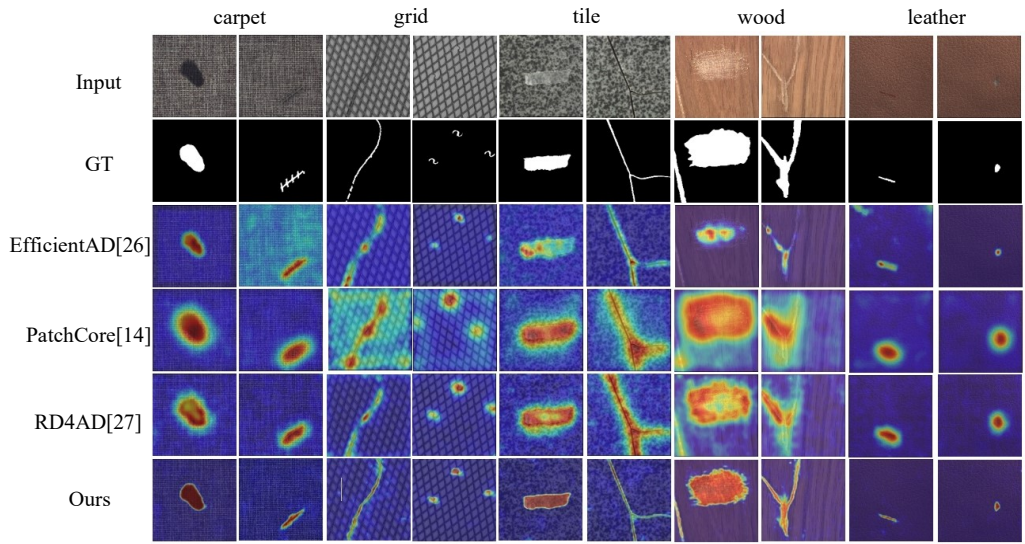

**Fig. 5.** The anomaly localization result of different methods on MVTec-AD textured images.

### 4.3 Experimental results

**Comparison with existing unsupervised methods.** Table 1 and Table 2 respectively show the detection and localization performance metrics of the proposed method and other anomaly detection methods on the MVTec-AD-Textured[11] dataset. We conduct a quantitative analysis among the anomaly detection and localization performances of ALMRR and six advanced methods, including DRAEM[18], EfficientAD[26], RD4AD[27], DSR[28], SimpleNet[29] and PNI[30]. An intuitive visual comparison between ALMRR and EfficientAD[26], PatchCore[14], and RD4AD[27] is shown in Fig. 5. ALMRR also performs well on other texture datasets, such as the MT-Defect[24] and NanoTWICE[25](shown in Fig. 6). Our method achieves state-of-the-art performance in all detection and localization metrics, and it is more precise and detailed in visualization result.

Table 1. Image-level anomaly detection of different method

| Category | Image AUROC(%) | | | | | | |
|---|---|---|---|---|---|---|---|
| | DRAEM[18] (ICCV21) | EfficientAD[26] (WACV24) | RD4AD[27] (CVPR22) | DSR[28] (ECCV22) | SimpleNet[29] (CVPR23) | PNI[30] (ICCV23) | Ours |
| grid | 99.9 | **100** | **100** | **100** | 99.7 | 98.4 | **100** |
| leather | **100** | 98.2 | **100** | **100** | **100** | **100** | **100** |
| tile | 99.6 | **100** | 99.3 | **100** | 99.8 | **100** | **100** |
| carpet | 97.0 | 99.4 | 98.9 | **100** | 99.7 | **100** | 99.6 |
| wood | 99.1 | 99.6 | 99.2 | 96.3 | **100** | 99.6 | **100** |
| avg | 99.1 | 99.4 | 99.5 | 99.3 | 99.8 | 99.6 | **99.9** |

Table 2. Pixel-level anomaly localization of different method

| Category | Pixel AUROC/Pixel AP(%) | | | | | | |
|---|---|---|---|---|---|---|---|
| | DRAEM[18] (ICCV21) | EfficientAD[26] (WACV24) | RD4AD[27] (CVPR22) | DSR[28] (ECCV22) | SimpleNet[29] (CVPR23) | PNI[30] (ICCV23) | Ours |
| grid | **99.7**/65.7 | 99.3/36.5 | 99.3/-- | --/**68.0** | 98.8/-- | 99.2/-- | 99.5/61.3 |
| leather | 98.6/75.3 | 99.6/54.6 | 99.4/-- | --/62.5 | 99.2/-- | 99.6/-- | **99.8/75.6** |
| tile | 99.2/92.3 | 98.1/74.2 | 95.6/-- | --/93.9 | 97.0/-- | 98.4/-- | **99.4/94.4** |
| carpet | 95.5/53.5 | 96.1/58.3 | 98.9/-- | --/78.2 | 98.2/-- | **99.4**/-- | 98.6/76.6 |
| wood | 96.4/77.7 | 96.3/60.8 | 95.3/-- | --/68.4 | 94.5/-- | 97.0/-- | **98.3/83.8** |
| avg | 97.9/72.9 | 97.9/56.9 | 97.7/-- | --/74.2 | 97.5/-- | 98.7/-- | **99.1/78.3** |

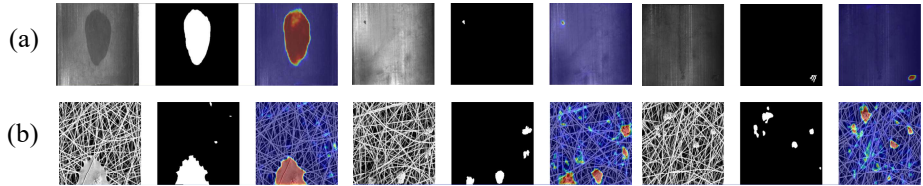

Fig. 6. The anomaly localization heatmaps on other real-world industrial datasets. (a) MT Defect dataset  (b) NanoTWICE dataset

Table 3. Quantitative comparison across different architectures

| Category | Pixel AUROC/Pixel AP(%) | | | |
|---|---|---|---|---|
| | CNN(1 × 1) | CNN(3 × 3) | Transformer | Mamba |
| grid | 99.4/58.3 | 99.3/56.9 | 99.4/54.4 | **99.5/61.3** |
| leather | 99.7/73.2 | 99.7/74.9 | **99.8**/74.2 | **99.8/75.6** |
| tile | 99.0/94.0 | 99.1/84.5 | **99.6**/93.7 | 99.4/**94.4** |
| carpet | 98.5/**78.8** | 97.8/75.6 | **98.6**/73.3 | **98.6**/76.6 |
| wood | 97.6/80.4 | 97.8/80.9 | 97.7/78.8 | **98.3/83.8** |
| avg | 98.8/76.9 | 98.7/74.6 | 99.0/74.9 | **99.1/78.3** |

Table 4. Comparison of latency time per image across different architectures

| | CNN(1 × 1) | CNN(3 × 3) | Transformer | Mamba |
|---|---|---|---|---|
| Time(s) | 0.10 | 0.13 | 0.18 | 0.12 |

## 4.4 Ablation Study

This subsection discusses the roles of the key modules in the proposed ALMRR on the MVTec-AD[11] textured images.

**Comparison of the Feature Reconstruction Module: Mamba-Based versus CNN and Transformer.** The Feature Reconstruction Module compared three different architectures: CNN (kernal = $1 \times 1$ / $3 \times 3$), Transformer, and Mamba. Table 3 presents the comparative results of the different Feature Reconstruction Modules. The experimental results indicate that the performance metrics of the three architectures do not differ significantly on pixel-AUROC. However, in terms of pixel-AP, Mamba is higher than CNN (1×1), CNN (3×3), and Transformer by 1.4%, 3.7%, and 3.4%, respectively. Comparison of latency time per image across different architectures, as shown in Table 4, demonstrates the low latency of the Mamba feature refinement module.

**Table 5.** Qualitative Comparison of the Advantages of the FRM

| Category | Image AUROC / Pixel AUROC / Pixel AP(%) | |
|---|---|---|
| | No Refinement Module | Refinement Module |
| grid | 92.0/98.6/55.3 | **100.0/99.5/61.3** |
| leather | 97.5/98.8/68.9 | **100.0/99.8/75.6** |
| tile | 98.5/99.0/90.4 | **100.0/99.4/94.4** |
| carpet | 96.9/98.0/73.1 | **99.6/98.6/76.6** |
| wood | 98.5/97.8/76.2 | **100.0/98.3/83.8** |
| avg | 96.7/98.4/72.8 | **99.9/99.1/78.3** |

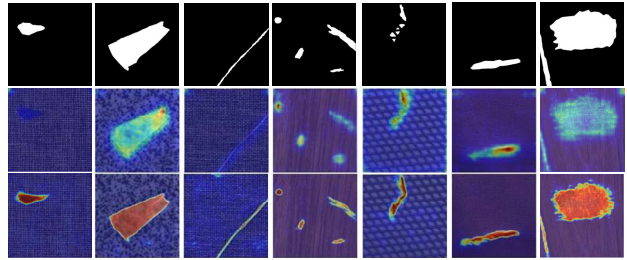

**Fig. 7.** Compared the effect of FRM (the first row is GT, the second row without using FRM, the third row with FRM).

**Validating the Effectiveness of the Feature Refinement Module.** We qualitatively and quantitatively compare the performance metrics before and after using the FRM (shown in Table 5) and the heatmaps (shown in Fig. 7). The experiment result convincingly demonstrates the effectiveness of our proposed FRM, which can significantly improve detection and localization accuracy.

**Effectiveness of Different Backbone.** The dual-branch feature extraction module with different architectural backbones is replaced, including VGG19, ResNet18, ResNet34, if the depth of feature is less than 1024, $1 \times 1$ convolutional layer is no longer necessary for dimensionality reduction. It has been verified that the model performs best on ResNet50 (shown in Table 6).

**Table 6.** Quantitative Comparison of Different Pre-trained Models

| Category | Pixel AUROC/Pixel AP(%) | | | |
|---|---|---|---|---|
| | VGG19 | ResNet18 | ResNet34 | ResNet50 |
| grid | 98.1/59.7 | 98.5/60.6 | 99.0/59.9 | **99.5**/**61.3** |
| leather | 98.8/74.2 | 99.4/74.4 | 99.5/75.0 | **99.8**/**75.6** |
| tile | 97.7/86.8 | 98.5/84.3 | 98.8/89.8 | **99.4**/**94.4** |
| carpet | 97.6/66.6 | 98.0/66.7 | 98.5/72.4 | **98.6**/**76.6** |
| wood | 96.0/75.2 | 96.9/75.7 | 97.4/76.5 | **98.3**/**83.8** |
| avg | 97.6/72.5 | 98.3/72.3 | 98.6/74.7 | **99.1**/**78.3** |

**Table 7.** Quantitative Comparison of Different blocks in ResNet50

| Category | Pixel AUROC/Pixel AP(%) | | | |
|---|---|---|---|---|
| | 1&2block | 2&3block | 1&2&3block | 1&2&3&4block |
| grid | 98.4/48.8 | 99.2/56.5 | **99.5/61.3** | 98.9/56.8 |
| leather | 99.2/65.8 | 99.6/68.5 | **99.8/75.6** | 99.6/73.8 |
| tile | 98.5/89.8 | 98.8/92.6 | **99.4**/94.4 | 99.0/**94.9** |
| carpet | 95.8/68.4 | 97.7/**77.2** | **98.6**/76.6 | 97.8/72.5 |
| wood | 97.5/74.9 | 98.3/80.4 | 98.3/83.8 | **99.2**/**84.5** |
| avg | 97.9/69.6 | 98.7/75.0 | **99.1**/**78.3** | 98.9/76.5 |

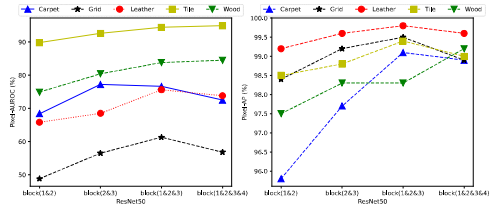

**Fig. 8.** The pixel-AUROC and pixel-AP metrics of our method on texture categories with different residual blocks of ResNet50.

**Effectiveness of Different ResNet50 Residual Block.** We compared the performance using different residual blocks of ResNet50. Experiments in Table 7 and Fig. 8 show that the best metrics are obtained by taking the features after the first three residual blocks. This also confirms that the abstract features from the last residual block, as mentioned earlier, do not enhance the effectiveness for texture images.

## 5 Conclusion

In this paper, we propose a novel anomaly localization method named ALMRR. During training stage, ALMRR initially employs a dual-branch structure to extract multi-scale features, followed by a Mamba feature reconstruction module that performs feature reconstruction based on normal sample information. Finally, the features are fed into the feature refinement module for detailed localization of anomalies. Our method achieves state-of-the-art results on MVTec-AD-Textured dataset. Ablation studies explored several key factors of our method and further demonstrated the effectiveness of our proposed MFRM and FRM strategies.

## Acknowledgments

This work is supported by the National Natural Science Foundation of China under Grant No. 62373350, the Youth Innovation Promotion Association CAS (2023145) and the Beijing Municipal Natural Science Foundation, China, under Grant L243018.

## References

1. Tao, X., Gong, X., Zhang, X., Yan, S., & Adak, C.: Deep learning for unsupervised anomaly localization in industrial images: A survey. IEEE Transactions on Instrumentation and Measurement, 71, 1-21 (2022)
2. Zhu, Lianghui, et al.: "Vision mamba: Efficient visual representation learning with bidirectional state space model." arXiv preprint arXiv:2401.09417 (2024)
3. Bergmann, P., Löwe, S., Fauser, M., Sattlegger, D., & Steger, C.: Improving unsupervised defect segmentation by applying structural similarity to autoencoders. arXiv preprint arXiv:1807.02011 (2018)
4. Tan, D. S., Chen, Y. C., Chen, T. P. C., & Chen, W. C.: Trustmae: A noise-resilient defect classification framework using memory-augmented auto-encoders with trust regions. In Proceedings of the IEEE/CVF winter conference on applications of computer vision. pp. 276-285 (2021)
5. Matsubara, T., Sato, K., Hama, K., Tachibana, R., & Uehara, K.: Deep generative model using unregularized score for anomaly detection with heterogeneous complexity. IEEE Transactions on Cybernetics, 52(6), 5161-5173 (2020)
6. Dehaene, D., & Eline, P.: Anomaly localization by modeling perceptual features. arXiv preprint arXiv:2008.05369. (2020)
7. Schlegl, T., Seeböck, P., Waldstein, S. M., Schmidt-Erfurth, U., & Langs, G.: Unsupervised anomaly detection with generative adversarial networks to guide marker discovery. In International conference on information processing in medical imaging. pp. 146-157 (2017)
8. Tang, T. W., Kuo, W. H., Lan, J. H., Ding, C. F., Hsu, H., & Young, H. T.: Anomaly detection neural network with dual auto-encoders GAN and its industrial inspection applications. Sensors, 20(12), 3336 (2020)
9. Dosovitskiy, Alexey, et al. "An image is worth 16x16 words: Transformers for image recognition at scale." arXiv preprint arXiv:2010.11929 (2020)
10. He, K., Zhang, X., Ren, S., & Sun, J.: Deep residual learning for image recognition. In Proceedings of the IEEE conference on computer vision and pattern recognition. pp. 770-778 (2016)
11. Bergmann, P., Fauser, M., Sattlegger, D., & Steger, C.: MVTec AD--A comprehensive real-world dataset for unsupervised anomaly detection. In Proceedings of the IEEE/CVF conference on computer vision and pattern recognition. pp. 9592-9600 (2019)
12. Deng, J., Dong, W., Socher, R., Li, L. J., Li, K., & Fei-Fei, L.: ImageNet: A large-scale hierarchical image database. In 2009 IEEE conference on computer vision and pattern recognition. pp. 248-255 (2009)
13. Defard, T., Setkov, A., Loesch, A., & Audigier, R.: Padim: a patch distribution modeling framework for anomaly detection and localization. In International Conference on Pattern Recognition. pp. 475-489 (2021)

14. Roth, K., Pemula, L., Zepeda, J., Schölkopf, B., Brox, T., & Gehler, P.: Towards total recall in industrial anomaly detection. In Proceedings of the IEEE/CVF Conference on Computer Vision and Pattern Recognition. pp. 14318-14328 (2022)
15. Yang, M., Wu, P., & Feng, H.: MemSeg: A semi-supervised method for image surface defect detection using differences and commonalities. Engineering Applications of Artificial Intelligence, 119, 105835. (2023)
16. Yang, J., Shi, Y., & Qi, Z.: DFR: Deep feature reconstruction for unsupervised anomaly segmentation. arXiv preprint arXiv:2012.07122 (2020)
17. You, Z., Yang, K., Luo, W., Cui, L., Zheng, Y., & Le, X.: ADTR: Anomaly detection transformer with feature reconstruction. In International Conference on Neural Information Processing. pp. 298-310 (2022)
18. Zavrtanik, V., Kristan, M., & Skočaj, D.: DRAEM-a discriminatively trained reconstruction embedding for surface anomaly detection. In Proceedings of the IEEE/CVF International Conference on Computer Vision. pp. 8330-8339 (2021)
19. Zavrtanik, V., Kristan, M., & Skočaj, D.: Reconstruction by inpainting for visual anomaly detection. Pattern Recognition, 112, 107706 (2021)
20. Pirnay, J., & Chai, K.: Inpainting transformer for anomaly detection. In International Conference on Image Analysis and Processing. pp. 394-406 (2022)
21. Zhang, H., Wang, Z., Wu, Z., & Jiang, Y. G.: DiffusionAD: Denoising diffusion for anomaly detection. arXiv preprint arXiv:2303.08730 (2023)
22. Liu, T., Li, B., Du, X., Jiang, B., Geng, L., Wang, F., & Zhao, Z.: FAIR: Frequency-aware Image Restoration for Industrial Visual Anomaly Detection. arXiv preprint arXiv: 2309.07068 (2023)
23. Tao, X., Adak, C., Chun, P. J., Yan, S., & Liu, H.: ViTALnet: Anomaly on industrial textured surfaces with hybrid transformer. IEEE Transactions on Instrumentation and Measurement, 72, 1-13 (2023)
24. Huang, Y., Qiu, C., & Yuan, K.: Surface defect saliency of magnetic tile. The Visual Computer, 36(1), 85-96 (2020)
25. Carrera, D., Manganini, F., Boracchi, G., & Lanzarone, E.: Defect detection in SEM images of nanofibrous materials. IEEE Transactions on Industrial Informatics, 13(2), 551-561 (2016)
26. Batzner, K., Heckler, L., & König, R.: EfficientAD: Accurate visual anomaly detection at millisecond-level latencies. In Proceedings of the IEEE/CVF Winter Conference on Applications of Computer Vision. pp. 128-138 (2024)
27. Deng, H., & Li, X.: Anomaly detection via reverse distillation from one-class embedding. In Proceedings of the IEEE/CVF Conference on Computer Vision and Pattern Recognition pp. 9737-9746 (2022)
28. Zavrtanik, V., Kristan, M., & Skočaj, D.: DSR–a dual subspace re-projection network for surface anomaly detection. In European conference on computer vision. pp. 539-554 (2022)
29. Liu, Z., Zhou, Y., Xu, Y., & Wang, Z.: SimpleNet: A simple network for image anomaly detection and localization. In Proceedings of the IEEE/CVF Conference on Computer Vision and Pattern Recognition. pp. 20402-20411 (2023)
30. Bae, J., Lee, J. H., & Kim, S.: PNI: industrial anomaly detection using position and neighborhood information. In Proceedings of the IEEE/CVF International Conference on Computer Vision. pp. 6373-6383 (2023)
31. Gu, A., & Dao, T.: Mamba: Linear-time sequence modeling with selective state spaces. arXiv preprint arXiv:2312.00752 (2023)